%% file: acl_latex.tex
\documentclass[11pt]{article}

\usepackage[preprint]{acl}

\usepackage{times}
\usepackage{latexsym}

\usepackage[T1]{fontenc}

\usepackage[utf8]{inputenc}

\usepackage{microtype}

\usepackage{inconsolata}

\usepackage{graphicx}
\usepackage{amsmath,amssymb}
\usepackage{booktabs}
\usepackage{multirow}
\usepackage{xcolor}
\usepackage{float}
\usepackage{needspace}
\title{ReaDy: Recurrent States as Dynamic Context for Speculative Drafting in Hybrid Linear Attention Large Language Models}

\author{
  \textbf{Bohan Sun}\textsuperscript{1,2}\thanks{Bohan Sun and Zunhai Su contributed equally to this work.}\quad
  \textbf{Zunhai Su}\textsuperscript{1,3,*}\quad
  \textbf{Xialie Zhuang}\textsuperscript{1}\quad
  \textbf{Jianjin Li}\textsuperscript{1}\\[0.5ex]
  \textbf{Ngai Wong}\textsuperscript{3}\thanks{Ngai Wong and Chuan-Wei Kuo are co-corresponding authors.}\quad \textbf{Chuan-Wei Kuo}\textsuperscript{1}\footnotemark[2]\\
  \textsuperscript{1}StartLux\\
  \textsuperscript{2}University of Chinese Academy of Sciences\\
  \textsuperscript{3}The University of Hong Kong
}

\begin{document}\raggedbottom
\maketitle
\begin{abstract}

Speculative decoding accelerates autoregressive generation by drafting multiple tokens per step, yet feature-based drafters rely on target hidden features whose contextual emphasis may not align with the needs of subsequent draft positions. Hybrid linear attention (HLA) models maintain compact recurrent states that accumulate historical associations, offering an efficient source of complementary context for speculative drafting. We introduce ReaDy, the first speculative decoding method, to the best of our knowledge, to leverage HLA recurrent states for position-specific context retrieval during drafting. ReaDy equips each draft position with a learned query to retrieve supplementary context directly from the target's committed recurrent states. A lightweight state reader integrates this context into draft representations through residual updates, while low-rank state reads and fused operators keep additional computation and I/O overhead low. Building on DSpark's parallel backbone and sequential prediction head, ReaDy employs a lightweight two-stage adaptation strategy, with reader capacity, state sources, and residual placement examined through ablation studies. Across seven benchmarks spanning mathematics, code generation, and dialogue, ReaDy achieves an average 3.16$\times$ decoding speedup over autoregressive decoding with Qwen3.8-27B on a single NVIDIA B200 GPU. It outperforms DSpark on every benchmark, improving throughput by 7.97\% and acceptance length by 8.50\% on average. Our code is available at \url{https://github.com/StartLuxLabs/ReaDy}.

\end{abstract}

\section{Introduction}
\label{sec:introduction}

\begin{figure*}[t]
  \centering
  \includegraphics[width=\textwidth]{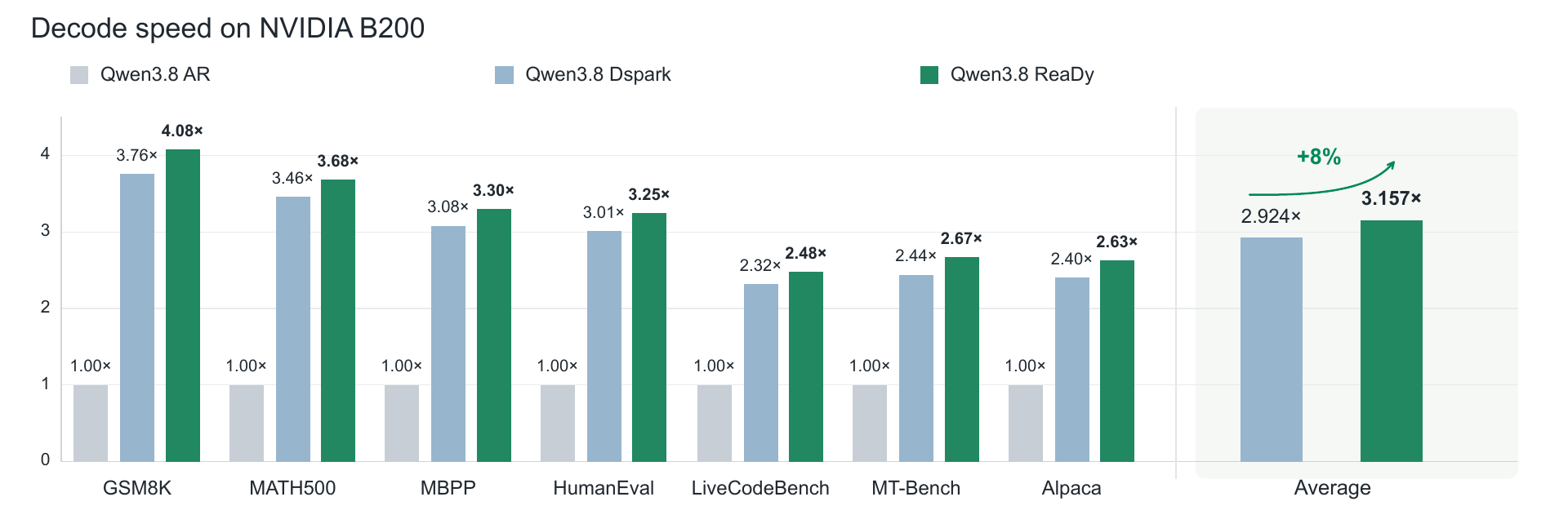}
  \caption{Decoding speedups of Qwen3.8-27B on a single NVIDIA B200 GPU at temperature zero. Each benchmark is normalized to autoregressive decoding. The Average panel reports the arithmetic mean of per-benchmark speedups. ReaDy improves throughput over DSpark across all seven benchmarks.}
  \label{fig:ready-speedup-overview}
\end{figure*}

Hybrid linear attention (HLA) offers a practical route to efficient long-context language modeling. By combining recurrent layers with full or sparse attention, HLA reduces the cost of maintaining context while retaining fine-grained retrieval capabilities~\citep{yang2025gateddeltanet,qwen2025next}. Autoregressive generation nevertheless remains sequential. Speculative decoding addresses this bottleneck: a lightweight drafter proposes several tokens for the target to verify together, reducing the number of target invocations needed to generate a sequence~\citep{leviathan2023speculative}.

Recent advances reduce the sequential cost of drafting itself. DFlash predicts token blocks in parallel from target hidden features~\citep{chen2026dflash}; DSpark combines a parallel backbone with a lightweight sequential head to capture dependencies within each block~\citep{cheng2026dspark}. Training that exposes drafters to their own predictions or accounts for verification further improves draft--target agreement~\citep{li2025eagle3,gu2026vat}. For semi-autoregressive drafting, however, the backbone must prepare representations for several future positions at once. Its predictions therefore depend on whether the supplied context supports the needs of the entire block.

Target hidden features provide a selective view of the prefix. Learned under next-token prediction, they reflect context selected by the target's queries at observed positions~\citep{liu2026kvshot}. Later draft positions can require relationships that those readouts do not emphasize. The drafter can attend over the supplied features, but this does not give it direct access to the underlying target memory. KVShot enables fresh retrieval through the target's KV cache~\citep{liu2026kvshot}. That access, however, reads token-wise memory whose size grows with the prefix, increasing the I/O cost of supplying additional context.

HLA offers a different source: recurrent states that compactly accumulate historical associations in fixed-size matrices. Direct access to these states lets each draft position seek supplementary context with its learned query, without scanning a growing token-wise cache. The challenge is to make this information useful for block prediction while keeping computation and I/O cost low.

We introduce \textbf{ReaDy}, a state-conditioned speculative decoding method for HLA models. ReaDy augments DSpark's parallel backbone with a reader that forms a query from each draft representation, reads selected target states, and injects the retrieved context through a residual update. It retains DSpark's sequential prediction head and target verification, allowing only committed states to inform subsequent drafting rounds. We summarize our contributions as follows:

\begin{itemize}
  \setlength{\itemsep}{0pt}
  \setlength{\parsep}{0pt}
  \setlength{\topsep}{0pt}
  \setlength{\partopsep}{0pt}
  \setlength{\baselineskip}{0.95\baselineskip}
  \item To the best of our knowledge, we are the first to propose leveraging HLA recurrent states as a complementary source of context for speculative drafting. Our theoretical analysis characterizes when new state queries can reveal information not determined by existing readouts, motivating position-specific retrieval from committed states.
  \item We develop ReaDy, a recurrent-state-conditioned speculative decoding method that equips each draft position with a learned state query and integrates the retrieved context through lightweight residual updates. Low-rank state reads, fused operators, and a two-stage adaptation strategy enable efficient integration with a pretrained drafter.
  \item Experiments across seven benchmarks with Qwen3.8-27B demonstrate that ReaDy improves decoding throughput and acceptance length over DSpark by \textbf{7.97\%} and \textbf{8.50\%} on average, respectively, achieving an average \textbf{3.16$\times$} speedup over autoregressive decoding.
\end{itemize}

\input{ready-architecture-figure}\input{ready-theoretical-motivation}\input{ready-method}\begin{table*}[!t]
  \centering
  \caption{Decoding performance on a single NVIDIA B200 GPU. TPS excludes prefill; $\times$AR is the speedup over autoregressive decoding. The final column reports ReaDy's TPS gain over DSpark. Acceptance length $\tau$ is the number of generated tokens per verification step, including target-supplied tokens. The average row uses mean TPS and mean $\tau$; speedups are computed from mean TPS. Best results are bold.}
  \label{tab:main-results}
  \small
  \setlength{\tabcolsep}{3.5pt}
  \setlength{\aboverulesep}{1.5pt}
  \setlength{\belowrulesep}{2pt}
  \renewcommand{\arraystretch}{1.0}
  \begin{tabular*}{\linewidth}{@{\extracolsep{\fill}}llrrrrrrrr@{}}
    \toprule
    \multicolumn{10}{@{}l}{\textbf{Qwen3.8-27B}\hspace{1.5em}Temperature = 0\hspace{1.5em}Draft block size = 7} \\
    \midrule
    \multirow{2}{*}{Task} & \multirow{2}{*}{Benchmark}
      & AR & \multicolumn{3}{c}{DSpark} & \multicolumn{4}{c}{\textbf{ReaDy (ours)}} \\
    \cmidrule(lr){3-3}\cmidrule(lr){4-6}\cmidrule(l){7-10}
    & & TPS & TPS & $\times$AR & $\tau$ & TPS & $\times$AR & $\tau$ & vs.\ DSpark \\
    \midrule
    \multirow{2}{*}{Math} & GSM8K & 96.40 & 362.25 & 3.76$\times$ & 5.28 & \textbf{393.60} & \textbf{4.08}$\boldsymbol{\times}$ & \textbf{5.77} & \textbf{+8.66\%} \\
     & MATH-500 & 95.99 & 331.90 & 3.46$\times$ & 4.93 & \textbf{353.23} & \textbf{3.68}$\boldsymbol{\times}$ & \textbf{5.27} & \textbf{+6.43\%} \\
    \addlinespace[1.5pt]
    \multirow{3}{*}{Code} & HumanEval & 95.60 & 287.95 & 3.01$\times$ & 4.34 & \textbf{310.29} & \textbf{3.25}$\boldsymbol{\times}$ & \textbf{4.70} & \textbf{+7.76\%} \\
     & MBPP & 96.21 & 296.00 & 3.08$\times$ & 4.34 & \textbf{317.55} & \textbf{3.30}$\boldsymbol{\times}$ & \textbf{4.68} & \textbf{+7.28\%} \\
     & LiveCodeBench & 94.33 & 219.23 & 2.32$\times$ & 3.49 & \textbf{234.28} & \textbf{2.48}$\boldsymbol{\times}$ & \textbf{3.75} & \textbf{+6.86\%} \\
    \addlinespace[1.5pt]
    \multirow{2}{*}{Chat} & MT-Bench & 96.11 & 234.54 & 2.44$\times$ & 3.47 & \textbf{256.71} & \textbf{2.67}$\boldsymbol{\times}$ & \textbf{3.81} & \textbf{+9.45\%} \\
     & Alpaca & 96.38 & 231.09 & 2.40$\times$ & 3.37 & \textbf{253.79} & \textbf{2.63}$\boldsymbol{\times}$ & \textbf{3.73} & \textbf{+9.83\%} \\
    \midrule
    \multicolumn{2}{@{}l}{\textbf{Average}} & 95.86 & 280.42 & 2.93$\times$ & 4.17 & \textbf{302.78} & \textbf{3.16}$\boldsymbol{\times}$ & \textbf{4.53} & \textbf{+7.97\%} \\
    \bottomrule
  \end{tabular*}
\end{table*}
\input{ready-position-figure}

\section{Experiments}
\label{sec:experiments}

We evaluate ReaDy's decoding performance and examine how its state-reading architecture balances draft acceptance against execution cost.

\subsection{Training and Evaluation Setup}
\label{sec:experimental-setup}

\input{ready-training}\paragraph{Inference settings.}
We compare ReaDy with autoregressive (AR) decoding and the pretrained DSpark drafter. Each model instance runs on one NVIDIA B200 GPU with SGLang~\citep{zheng2024sglang}, using BF16 without quantization and a request concurrency of one. Evaluation requests are distributed across GPUs, with each input compared across methods on the same GPU. All methods use greedy decoding (temperature zero), with thinking enabled, a maximum of 1,024 output tokens, and a context limit of 16,384 tokens. Generation stops upon EOS. Both DSpark and ReaDy use a draft block size of seven. Unless otherwise specified, ReaDy reads target states from layers 30 and 62 with reader rank eight and injects the resulting residual once, after attention in the final draft layer.

\paragraph{Benchmarks and metrics.}
We evaluate seven benchmarks across three task categories. Mathematical reasoning is evaluated on GSM8K~\citep{cobbe2021gsm8k} and MATH500~\citep{hendrycks2021math}. For code generation, we use sanitized MBPP~\citep{austin2021mbpp}, HumanEval~\citep{chen2021humaneval}, and LiveCodeBench~\citep{jain2024livecodebench}. Dialogue and instruction following are evaluated using the first turn of MT-Bench~\citep{zheng2023mtbench} and a fixed Alpaca subset~\citep{taori2023alpaca}. We report server-side decoding throughput in tokens per second (TPS), excluding prefill. Acceptance length is the total number of output tokens divided by the total number of verification rounds, including tokens supplied by the target model; it therefore measures the average progress per verification. For the overall throughput comparison, we first average TPS across benchmarks and then compute speedups from these averages.

\subsection{Main Results}
\label{sec:main-results}

Table~\ref{tab:main-results} shows that ReaDy improves both throughput and acceptance length on every benchmark. Its mean decoding throughput reaches 302.78 tokens/s, compared with 95.86 for AR decoding and 280.42 for DSpark, yielding a 3.16$\times$ speedup over AR and a further 7.97\% gain over DSpark. The improvement extends across task types: throughput gains over DSpark range from 6.43\% to 8.66\% on mathematics and code generation, and reach 9.45\% on MT-Bench and 9.83\% on Alpaca. The corresponding overall acceptance-length gain is 8.50\%. Thus, the additional progress per verification is sufficient to offset the cost of state retrieval in the evaluated setting. The slightly smaller throughput gain also highlights the need to control reader overhead, which motivates the design comparisons below. Appendix~\ref{app:qualitative-example} illustrates how an early mismatch affects block acceptance in an individual generation.

\input{ready-position-analysis}\input{ready-ablations}\input{ready-related-work-and-conclusion}

\bibliography{ready-references,ready-introduction,ready-related-work,ready-background}

\clearpage\appendix\setcounter{equation}{0}\renewcommand{\theequation}{A.\arabic{equation}}\input{ready-theory-appendix}\input{ready-training-appendix}\input{ready-qualitative-figure}\input{ready-qualitative-example}\input{ready-background-appendix}\input{ready-position-protocol}\end{document}

%% file: ready-architecture-figure.tex
\begin{figure*}[!t]
  \centering
  \includegraphics[width=\textwidth]{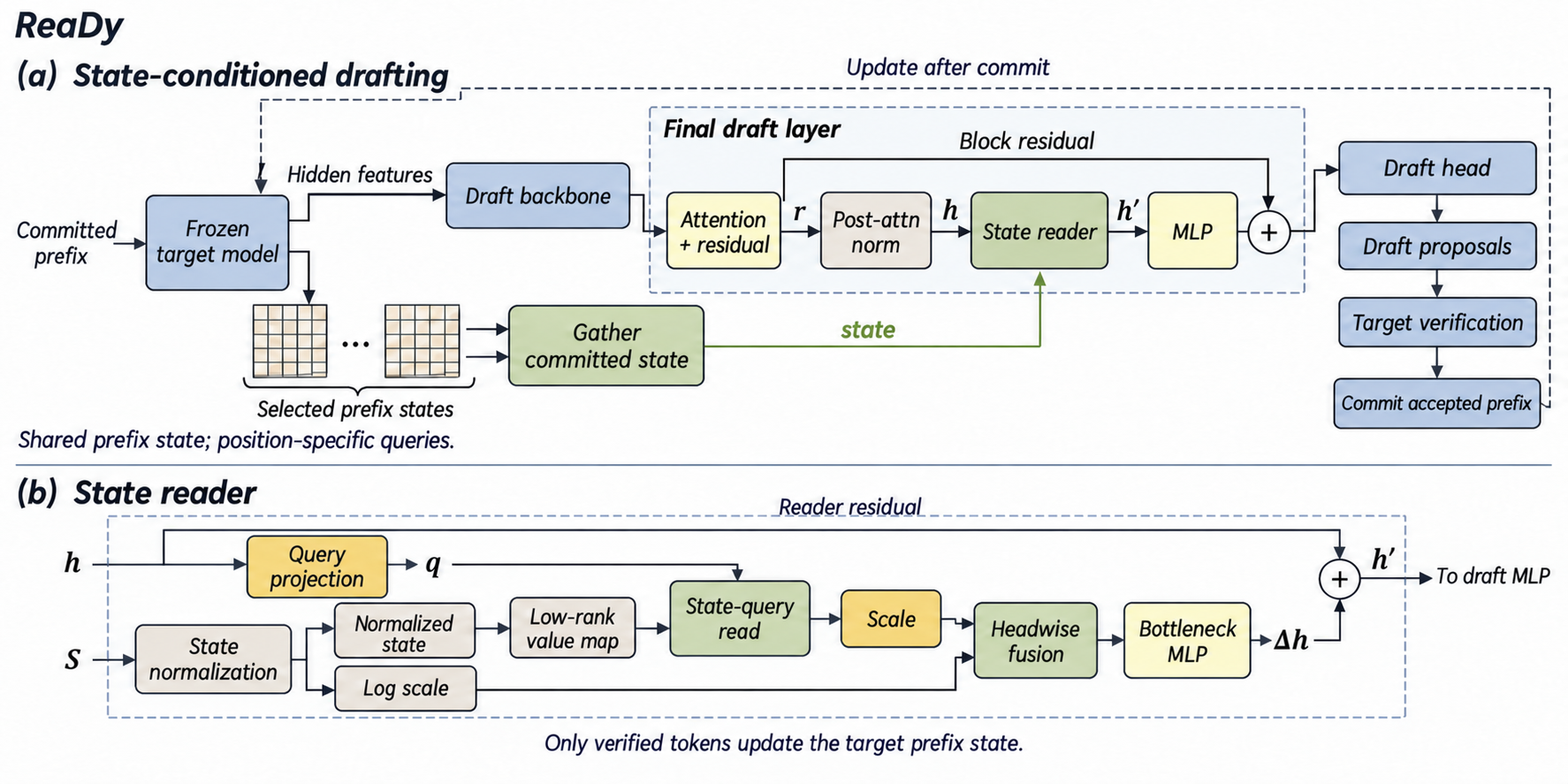}
  \caption{ReaDy's state-conditioned drafting and verification process. (a) The drafter combines target hidden features with position-specific reads from committed recurrent states. Verification determines the accepted prefix and the states available in the next round. (b) Query projection maps each draft representation through a rank-8 bottleneck to the 128-dimensional key space. The value map compresses the state values to rank eight. Position-specific readouts are fused with state-scale information and injected through a residual connection.}
  \label{fig:ready-architecture}
\end{figure*}

%% file: ready-theoretical-motivation.tex
\section{Theoretical Motivation}
\label{sec:theoretical-motivation}

ReaDy builds on the distinction between a recurrent state and a query-specific readout of that state. We analyze this distinction at the attention-head level to motivate an additional route to prefix context. Full proofs and assumptions are given in Appendix~\ref{app:state-readout-analysis}; Appendix~\ref{app:backbone-background} reviews the underlying attention and speculative-decoding mechanisms.

\paragraph{Query-dependent context.}
Target queries are learned under next-token prediction, and their readouts reflect position-specific uses of context~\citep{liu2026kvshot}. Their contextual focus need not match the needs of later positions in a draft block. To isolate the selection, consider an idealized interface exposing readouts $O=SQ$ of a fixed state $S\in\mathbb{R}^{d_v\times d_k}$, with known queries $Q=[q_1,\ldots,q_m]$. Let $\Pi_Q=QQ^\dagger$ project onto their span, where $Q^\dagger$ is the Moore--Penrose pseudoinverse. A new query $q_*$ yields
\begin{equation}
  Sq_*=(SQ)Q^\dagger q_*+S(I-\Pi_Q)q_*.
  \label{eq:query-decomposition}
\end{equation}
The first term is the response to the component of $q_*$ within the exposed query span and is determined by the existing readouts. The second captures the response along directions outside that span. For unrestricted $S$, when this component of $q_*$ is nonzero, different states can agree on every exposed readout while producing different answers to $q_*$. Conversely, if $Q$ has full row rank, its span covers the key space and no such ambiguity remains. The relevant issue is therefore which query directions the interface exposes, not the number of readouts alone. This idealized result concerns state-query readouts; it does not establish that the drafter's complete multilayer features are insufficient.

\paragraph{Revisiting the committed state.}
Gated DeltaNet updates its state as $S_i=S_{i-1}A_i+B_i$, where $A_i$ transforms existing associations and $B_i$ writes new ones, with raw attention readout $o_i=S_iq_i$~\citep{yang2025gateddeltanet}. Along a fixed target-model continuation, define $P_j=A_{t+1}\cdots A_{t+j}$ and let $W_j$ collect subsequent writes. Unrolling the recurrence gives
\begin{equation}
  \begin{aligned}
    S_{t+j} &= S_tP_j+W_j,\\
    o_{t+j} &= S_t\underbrace{(P_jq_{t+j})}_{q_j^{\mathrm{eff}}}+W_jq_{t+j}.
  \end{aligned}
  \label{eq:future-state-readout}
\end{equation}
The first term is the contribution of the committed prefix; the second collects the effects of future writes. The accumulated transitions change how prefix information contributes, but this contribution remains expressible as a new query to $S_t$. The effective query depends on the continuation and is not available during drafting; the identity motivates learning useful query directions from draft representations. Reading $S_t$ alone does not recover the second term, which depends on future writes.

These observations motivate ReaDy's context-access design. A learned reader maps each draft representation to a query of the same committed state, allowing the retrieved context to vary across the block without advancing the target state. Residual integration retains the backbone's role in modeling the continuation and combining that context with hidden features. The fixed-size state supplies this information without a token-wise memory scan. Section~\ref{sec:method} places these reads within ReaDy's semi-autoregressive drafting and verification process.

%% file: ready-method.tex
\section{Method}
\label{sec:method}

ReaDy is a DSpark-based, state-conditioned speculative decoding method for HLA models. It combines block prediction with position-conditioned reads from the target's committed recurrent states (Figure~\ref{fig:ready-architecture}). We use DSpark's parallel backbone and sequential prediction head~\citep{cheng2026dspark}, and introduce a state-reading path in the final draft layer. Section~\ref{sec:experimental-setup} describes initialization and training.

\input{ready-io-figure}

\subsection{State-Conditioned Block Drafting}
\label{sec:state-context}

\paragraph{State interface.}
At a committed prefix of length $t$, the target provides its hidden features and recurrent states from selected linear-attention layers. We write the state of head $a$ in layer $\ell$ as $S_t^{\ell,a}\in\mathbb{R}^{d_v\times d_k}$, using the value-by-key convention of Section~\ref{sec:theoretical-motivation}. These matrices summarize prefix associations without retaining a separate entry for every token. All draft positions access the same committed states; the information retrieved from them depends on the representation at each position.

\paragraph{Position-conditioned reading.}
We place the reader in the final draft layer, after post-attention normalization and before the MLP. At this point, the draft representation has already incorporated context through the backbone's existing attention path. Let $h_j\in\mathbb{R}^{d}$ denote this representation at draft position $j$, and let $\bar S$ be an RMS-normalized state. Suppressing layer and head indices, we express the read as
\begin{equation}
  \begin{aligned}
    z_j &= W_q\operatorname{RMSNorm}(h_j), \\
    q_j &= U^{\top}z_j, \qquad T=V^{\top}\bar S, \\
    u_j &= \frac{Tq_j}{\sqrt{d_k}},
  \end{aligned}
  \label{eq:state-read}
\end{equation}
Here $W_q\in\mathbb{R}^{r\times d}$ and $U\in\mathbb{R}^{r\times d_k}$ first form a rank-$r$ query bottleneck, then expand it to the full key space. The query $q_j\in\mathbb{R}^{d_k}$ is thus restricted to a learned subspace. Independently, $V\in\mathbb{R}^{d_v\times r}$ compresses the state's value dimension, producing $T\in\mathbb{R}^{r\times d_k}$ and an $r$-dimensional readout $u_j$. In our final configuration, $d_k=d_v=128$ and $r=8$: the reader forms an $8\times128$ projected state while retaining the 128-dimensional key axis. Both constraints apply to the reader; the target continues to maintain its original $128\times128$ recurrent states.

\paragraph{Fusion and residual injection.}
We collect the readouts from the selected layers and heads and concatenate them with their corresponding log-RMS state scales to form $f_j$. Normalization controls the scale of the state entries, while the appended log-RMS features retain information about their original magnitudes. A bottleneck MLP maps the fused features back to the draft width:
\begin{equation}
  \begin{aligned}
    \Delta h_j &= W_{\mathrm{up}}\operatorname{SiLU}\!\left(
      W_{\mathrm{down}}\operatorname{RMSNorm}(f_j)\right), \\
    h'_j &= h_j + \Delta h_j.
  \end{aligned}
  \label{eq:reader-residual}
\end{equation}
The draft MLP processes $h'_j$ to produce the features used for base logits. DSpark's sequential prediction head combines these logits with transition biases conditioned on the preceding token to generate the block. The residual combines retrieved prefix context with the backbone's features, leaving continuation modeling to the drafter. Section~\ref{sec:ablations} evaluates reader capacity, source layers, and injection placement.

\paragraph{Drafting, verification, and commitment.}
Each round uses target hidden features and recurrent states associated with the committed history. The parallel backbone forms the draft representations, the reader supplies position-specific context, and the sequential head produces the candidate block. Under our greedy decoding setting, the target accepts the longest candidate prefix matching its argmax predictions. The first mismatch ends acceptance: the remaining proposals are discarded, and a target-supplied correction is committed with the accepted prefix. If the whole block is accepted, the target supplies a bonus token, unless generation has terminated. Subsequent rounds read only states associated with committed tokens; states from rejected continuations are not retained. This process repeats until EOS or the output limit. ReaDy changes how proposals are conditioned while retaining the existing verification rule.

\subsection{Efficient State Access}
\label{sec:efficient-state-access}

\paragraph{State-access footprint.}
For $M$ selected layer--head pairs, the reader's source states contain $M d_v d_k$ elements, independent of prefix length $t$. A token-wise KV memory with the same head count and key/value widths contains $M t(d_k+d_v)$ elements. This comparison concerns the context-source footprint; actual memory traffic also depends on execution and reuse. The state-side projection $T$ is independent of draft position, so it can be shared across the queries in a block. The projected states contain $M r d_k$ elements, reducing the value width of each subsequent read from $d_v$ to $r$. Low rank reduces this read cost without eliminating the full-state access needed for preparation. This analysis concerns only the added context channel; the target and drafter retain their existing attention caches and computation.

\paragraph{Avoiding redundant data movement.}
A fixed-size context source can still incur substantial overhead if gathering and preparation materialize intermediate tensors. We optimize this path as illustrated in Figure~\ref{fig:ready-io-optimization}. A fused gather kernel writes selected committed states directly into a preallocated execution buffer, eliminating an intermediate tensor and an additional copy. We fuse pointwise state preparation and readout feature packing to avoid separate passes over intermediate results. Native RMSNorm kernels further avoid redundant casts. These changes reduce the execution cost of the reader without changing which committed states it accesses.

\begin{figure}[!htb]
  \centering
  \includegraphics[width=\columnwidth]{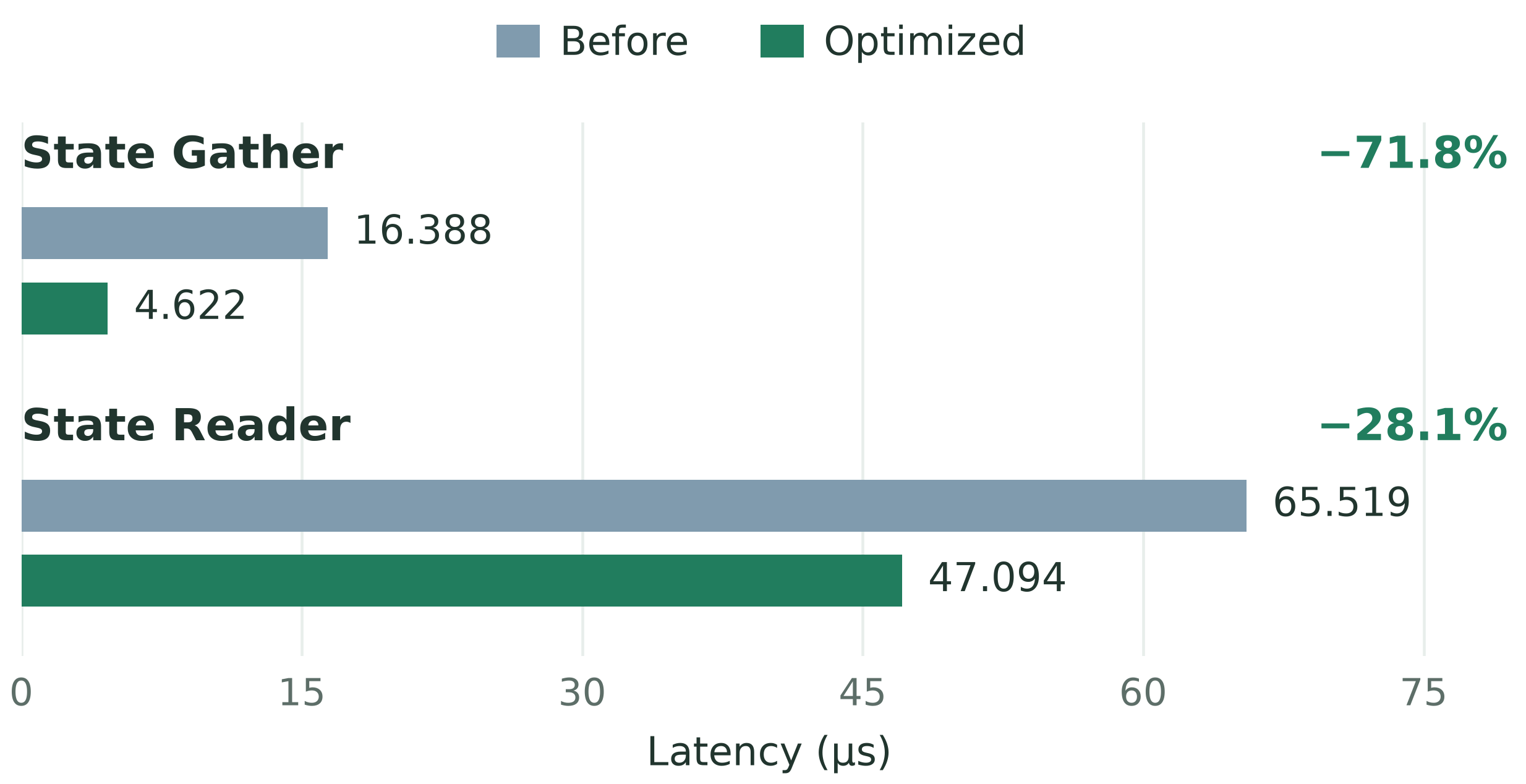}
  \caption{Operator latency before and after optimization. State gathering and state reading achieve latency reductions of 71.8\% and 28.1\%, respectively.}
  \label{fig:ready-operator-latency}
\end{figure}

As shown in Figure~\ref{fig:ready-operator-latency}, these optimizations reduce state-gather latency from 16.388 to 4.622~$\mu$s (71.8\%) and state-reader latency from 65.519 to 47.094~$\mu$s (28.1\%). These measurements quantify operator-level savings; end-to-end decoding gains are reported in Section~\ref{sec:main-results}.

%% file: ready-io-figure.tex
\begin{figure*}[!t]
  \centering
  \includegraphics[width=\textwidth]{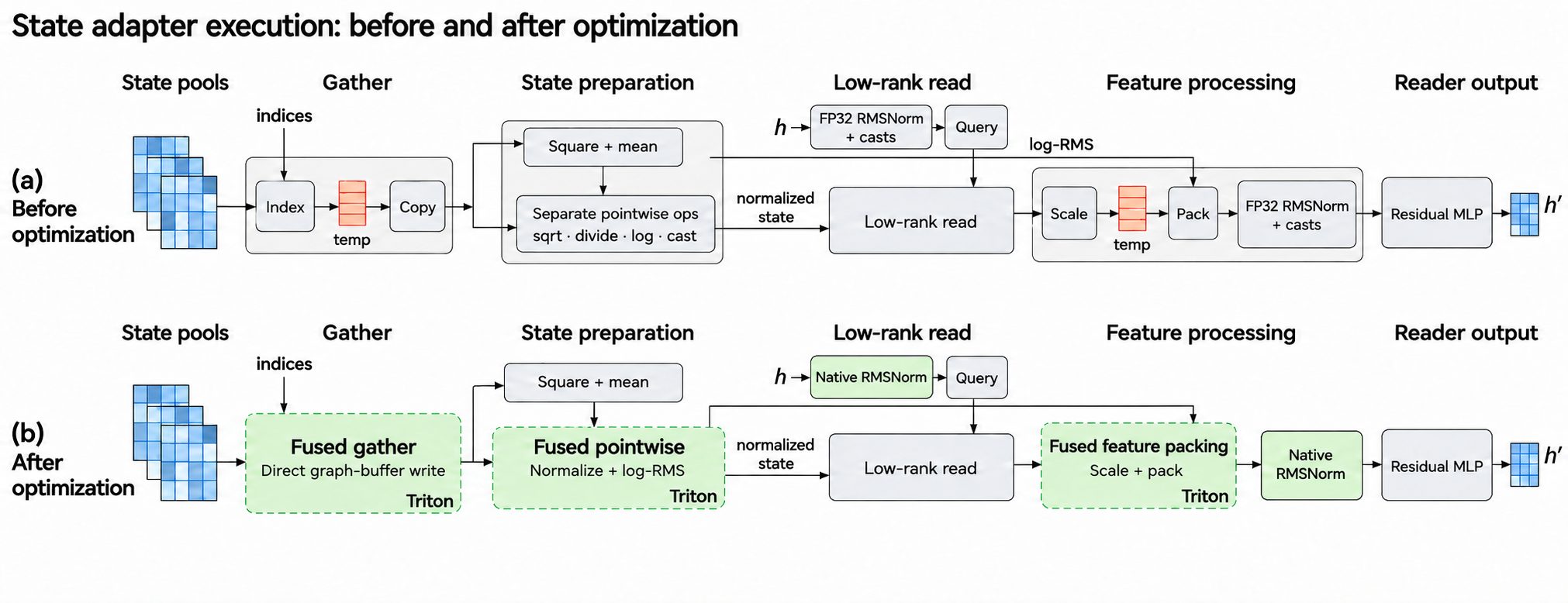}
  \caption{State-reader execution before and after optimization. Direct gathering and fused state preparation and feature packing reduce intermediate tensor materialization and redundant data movement. Green blocks indicate optimized operations.}
  \label{fig:ready-io-optimization}
\end{figure*}

%% file: ready-position-figure.tex
\begin{figure*}[!t]
  \centering
  \includegraphics[width=\textwidth]{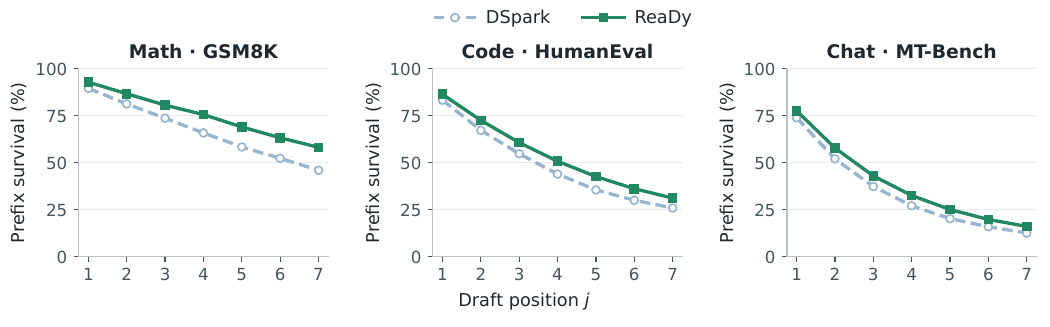}
  \caption{Draft-prefix survival in mathematics, code, and dialogue. At position $j$, the rate is the fraction of fully observable blocks in which all first $j$ proposals are accepted, pooled within each benchmark. ReaDy improves survival at every position. The curves show cumulative survival, not acceptance conditional on earlier positions.}
  \label{fig:ready-prefix-survival}
\end{figure*}

%% file: ready-training.tex
\paragraph{Initialization and two-stage training.}
We use Qwen3.8-27B as the target and initialize ReaDy's drafting backbone and prediction head from a pretrained DSpark model~\citep{cheng2026dspark} released by RadixArk.\footnote{\url{https://huggingface.co/RadixArk/Qwen3.8-27B-DSpark}} We adapt the drafter on Open-PerfectBlend.\footnote{\url{https://huggingface.co/datasets/mlabonne/open-perfectblend}} In the first stage, we freeze the drafter and train only the state reader on 20k samples. In the second stage, we jointly train the reader and the drafter's LoRA parameters~\citep{hu2021lora} on 50k samples. The target model remains frozen throughout. Both stages use
\begin{equation}
  \mathcal{L}=\mathcal{L}_{\mathrm{CE}}+0.1\,\mathcal{L}_{\mathrm{L1}},
  \label{eq:adaptation-loss}
\end{equation}
combining cross-entropy with an L1 loss between draft and target probability distributions, without a confidence loss. This objective and adaptation recipe differ from DSpark's original training. Comparisons with the pretrained DSpark baseline therefore reflect both adaptation and state access; Table~\ref{tab:finetuning-control} additionally compares ReaDy with fine-tuning without state access. Appendix~\ref{app:training} describes the data mixture, loss aggregation, and optimizer settings.

%% file: ready-position-analysis.tex
\subsection{Draft-Prefix Survival}
\label{sec:position-analysis}

Mean acceptance length summarizes overall progress but does not show how agreement changes across a draft block. We therefore examine all seven proposed positions using the final checkpoint. Figure~\ref{fig:ready-prefix-survival} reports cumulative prefix survival for Math (GSM8K), Code (HumanEval), and Chat (MT-Bench). At position $j$, this is the fraction of eligible drafting rounds in which all of the first $j$ proposed tokens are accepted, rather than the acceptance probability conditional on earlier positions surviving. ReaDy has higher prefix survival at every position in all three benchmarks. At the seventh position, survival increases from 45.98\% to 58.17\% on GSM8K, from 25.89\% to 31.19\% on HumanEval, and from 12.59\% to 16.11\% on MT-Bench. Thus, a larger fraction of ReaDy's complete seven-token proposals survive verification. Appendix~\ref{app:position-protocol} describes the measurement protocol and its scope.

%% file: ready-ablations.tex
\subsection{Ablation Studies}
\label{sec:ablations}
\input{ready-finetuning-control}

Tables~\ref{tab:ablation-rank}--\ref{tab:ablation-injection-site} compare reader capacity, state source layers, residual injection layers, and injection position. Each architectural experiment varies one component of ReaDy and reports mean decoding throughput (TPS), acceptance length ($\tau$), and their relative gains ($\Delta$ TPS and $\Delta\tau$, in percent) over the pretrained DSpark baseline. Bold values identify the best result in each table.

\paragraph{Reader capacity.}
Table~\ref{tab:ablation-rank} compares ranks 8, 16, 32, and 128 in both key and value spaces. Increasing rank from eight to 128 raises mean acceptance length from 4.5285 to 4.5404, but reduces the throughput gain over DSpark from 7.97\% to 7.81\%. Rank eight retains most of the acceptance benefit and gives the highest measured throughput, supporting a compact supplementary readout.

\begin{table}[H]
  \centering
  \caption{Reader rank. Full rank uses 128-dimensional key and value spaces.}
  \label{tab:ablation-rank}
  \small
  \setlength{\tabcolsep}{3pt}
  \setlength{\aboverulesep}{1.5pt}
  \setlength{\belowrulesep}{2pt}
  \renewcommand{\arraystretch}{1.0}
  \begin{tabular*}{\columnwidth}{@{\extracolsep{\fill}}lrrrr@{}}
    \toprule
    Reader rank & TPS & $\tau$ & $\Delta$ TPS & $\Delta\tau$ \\
    \midrule
    DSpark & 280.42 & 4.1739 & 0.00 & 0.00 \\
    \midrule
    8 & \textbf{302.78} & 4.5285 & \textbf{7.97} & 8.50 \\
    16 & 302.36 & 4.5298 & 7.82 & 8.53 \\
    32 & 302.21 & 4.5323 & 7.77 & 8.59 \\
    128 (full) & 302.31 & \textbf{4.5404} & 7.81 & \textbf{8.78} \\
    \bottomrule
  \end{tabular*}
\end{table}

\paragraph{State source layers.}
State usefulness varies substantially by source layer (Table~\ref{tab:ablation-state-layers}). Layer 62 alone yields a 7.48\% acceptance gain, compared with 2.45\% and 2.85\% for layers 2 and 30. Combining layers 30 and 62 raises acceptance and throughput gains to 8.50\% and 7.97\%; adding layer 2 reduces them to 6.89\% and 6.22\%. The two-layer combination therefore provides the best measured trade-off, suggesting complementary context without the cost of an additional state source.

\begin{table}[H]
  \centering
  \caption{State source layers (target layer IDs).}
  \label{tab:ablation-state-layers}
  \small
  \setlength{\tabcolsep}{3pt}
  \setlength{\aboverulesep}{1.5pt}
  \setlength{\belowrulesep}{2pt}
  \renewcommand{\arraystretch}{1.0}
  \begin{tabular*}{\columnwidth}{@{\extracolsep{\fill}}lrrrr@{}}
    \toprule
    Target layer IDs & TPS & $\tau$ & $\Delta$ TPS & $\Delta\tau$ \\
    \midrule
    DSpark & 280.42 & 4.1739 & 0.00 & 0.00 \\
    \midrule
    (2) & 286.11 & 4.2763 & 2.03 & 2.45 \\
    (30) & 287.17 & 4.2929 & 2.41 & 2.85 \\
    (62) & 300.21 & 4.4862 & 7.06 & 7.48 \\
    (30, 62) & \textbf{302.78} & \textbf{4.5285} & \textbf{7.97} & \textbf{8.50} \\
    (2, 30, 62) & 297.86 & 4.4614 & 6.22 & 6.89 \\
    \bottomrule
  \end{tabular*}
\end{table}

\paragraph{Number of injections.}
Table~\ref{tab:ablation-draft-layers} compares injecting context only in the final draft layer (layer 4), in layers 0, 2, and 4, and in all five layers. Mean acceptance lengths are 4.5285, 4.5288, and 4.5284, respectively, whereas throughput falls from 302.78 to 299.93 and 297.35 tokens/s as more readers are added. Additional readers reduce measured throughput without materially increasing the number of tokens generated per verification. A single injection in the final layer provides the best throughput in this comparison.

\begin{table}[H]
  \centering
  \caption{Injection layers (zero-indexed draft layer IDs).}
  \label{tab:ablation-draft-layers}
  \small
  \setlength{\tabcolsep}{3pt}
  \setlength{\aboverulesep}{1.5pt}
  \setlength{\belowrulesep}{2pt}
  \renewcommand{\arraystretch}{1.0}
  \begin{tabular*}{\columnwidth}{@{\extracolsep{\fill}}lrrrr@{}}
    \toprule
    Draft layer IDs & TPS & $\tau$ & $\Delta$ TPS & $\Delta\tau$ \\
    \midrule
    DSpark & 280.42 & 4.1739 & 0.00 & 0.00 \\
    \midrule
    (4) & \textbf{302.78} & 4.5285 & \textbf{7.97} & \textbf{8.50} \\
    (0, 2, 4) & 299.93 & \textbf{4.5288} & 6.96 & \textbf{8.50} \\
    (0, 1, 2, 3, 4) & 297.35 & 4.5284 & 6.04 & 8.49 \\
    \bottomrule
  \end{tabular*}
\end{table}

\Needspace{18\baselineskip}
\paragraph{Injection position.}
Within the final draft layer (Table~\ref{tab:ablation-injection-site}), injecting after attention improves mean acceptance length from 4.4993 to 4.5285 and throughput from 301.94 to 302.78 tokens/s relative to injecting before attention. The advantage is modest but favors the placement used in ReaDy. At this position, the query is formed after the draft representation has incorporated attention context, and the retrieved information can subsequently be processed by the MLP.

\begin{table}[H]
  \centering
  \caption{Injection position in the final draft layer.}
  \label{tab:ablation-injection-site}
  \small
  \setlength{\tabcolsep}{3pt}
  \setlength{\aboverulesep}{1.5pt}
  \setlength{\belowrulesep}{2pt}
  \renewcommand{\arraystretch}{1.0}
  \begin{tabular*}{\columnwidth}{@{\extracolsep{\fill}}lrrrr@{}}
    \toprule
    Injection site & TPS & $\tau$ & $\Delta$ TPS & $\Delta\tau$ \\
    \midrule
    DSpark & 280.42 & 4.1739 & 0.00 & 0.00 \\
    \midrule
    Pre-attention & 301.94 & 4.4993 & 7.68 & 7.80 \\
    Post-attention & \textbf{302.78} & \textbf{4.5285} & \textbf{7.97} & \textbf{8.50} \\
    \bottomrule
  \end{tabular*}
\end{table}

%% file: ready-finetuning-control.tex
\paragraph{Fine-tuning without state access.}
\label{sec:finetuning-control}
To assess the benefit beyond additional training, Table~\ref{tab:finetuning-control} includes a fine-tuned DSpark variant without the state adapter. Fine-tuning alone raises throughput from 280.42 to 286.50 tokens/s and acceptance length from 4.1739 to 4.2732, confirming that adaptation contributes to the improvement over the original checkpoint. ReaDy further reaches 302.78 tokens/s and an acceptance length of 4.5285: gains of 5.68\% and 5.97\%, respectively, over this no-state variant. This comparison supports an additional benefit from state-conditioned adaptation. The 7.97\% throughput gain over pretrained DSpark therefore reflects both continued adaptation and state access.

\begin{table}[H]
  \centering
  \caption{Fine-tuning control. FT denotes fine-tuning without state access; gains are relative to DSpark.}
  \label{tab:finetuning-control}
  \small
  \setlength{\tabcolsep}{3pt}
  \setlength{\aboverulesep}{1.5pt}
  \setlength{\belowrulesep}{2pt}
  \renewcommand{\arraystretch}{1.0}
  \begin{tabular*}{\columnwidth}{@{\extracolsep{\fill}}lrrr@{}}
    \toprule
    Method & TPS & $\tau$ & $\Delta$ TPS (\%) \\
    \midrule
    DSpark & 280.42 & 4.1739 & 0.00 \\
    DSpark + FT & 286.50 & 4.2732 & 2.17 \\
    ReaDy & \textbf{302.78} & \textbf{4.5285} & \textbf{7.97} \\
    \bottomrule
  \end{tabular*}
\end{table}

%% file: ready-related-work-and-conclusion.tex
\section{Related Work}
\label{sec:related-work}

\paragraph{Target context reuse for speculative drafting.}
Drafters can reuse target hidden features or query target memory directly. GliDe~\citep{du2024glide} and LongSpec~\citep{yang2026longspec} access the target's KV cache through cross-attention. KVShot~\citep{liu2026kvshot} compares hidden-feature and KV-cache reuse, highlighting the trade-off between richer context and retrieval cost. Although cache sharing avoids duplicate storage, reading token-wise history incurs memory traffic that grows with the number of attended tokens. ReaDy instead queries fixed-size recurrent states, using low-rank reads and fused operators to limit the additional I/O cost.

\paragraph{Speculative decoding for recurrent and hybrid models.}
Recurrent targets require dependency-aware verification and recovery of accepted states. \citet{wang2024mamballama} address these challenges for Mamba and hybrid models; STree~\citep{wu2025stree} extends tree verification to hybrid state-space models. SpecLA~\citep{wang2026specla} combines topology-aware verification and factor-based state recovery with EAGLE-style drafting from recurrent target features. ReaDy focuses on position-conditioned reads of committed states for block drafting. We analyze this context interface and study reader capacity, state sources, and residual placement within a DSpark-based implementation.

\section{Conclusion}
\label{sec:conclusion}

We presented ReaDy, a DSpark-based, state-conditioned speculative decoding method for HLA models. Its reader retrieves position-specific context from committed recurrent states and integrates it with draft representations through residual updates. Low-rank reads and fused operators control the additional I/O overhead. ReaDy achieves an average 3.16$\times$ speedup over autoregressive decoding and a 7.97\% throughput improvement over DSpark across seven benchmarks on Qwen3.8-27B using a single NVIDIA B200 GPU. These results demonstrate the value of recurrent-state retrieval for improving draft agreement and decoding efficiency within DSpark's semi-autoregressive architecture.

Future work will explore training the complete drafter from scratch and learning to allocate state access and readout capacity across draft positions.

%% file: ready-theory-appendix.tex
\section{Analysis of Recurrent-State Readouts}
\label{app:state-readout-analysis}

This appendix establishes the two statements used in Section~\ref{sec:theoretical-motivation}. We use value-by-key states and omit layer and head indices.

\subsection{Information Available from Existing Queries}
\label{app:query-identifiability}

Let $S$ be a real $d_v\times d_k$ matrix, $Q$ a known real $d_k\times m$ query matrix, and $O=SQ$. Set $\Pi_Q=QQ^\dagger$, the orthogonal projector onto the column space of $Q$.

\paragraph{Proposition 1.}
For an arbitrary query $q_*$ and unrestricted $S$, the readout $Sq_*$ is uniquely determined by $(O,Q)$ if and only if $q_*$ belongs to the column space of $Q$.

\paragraph{Proof.}
Splitting $q_*$ into its components within and orthogonal to the query span gives
\begin{equation}
  \begin{aligned}
    Sq_* &= S\Pi_Qq_*+S(I-\Pi_Q)q_*\\
         &= OQ^\dagger q_*+S(I-\Pi_Q)q_*.
  \end{aligned}
\end{equation}
If $q_*$ lies in the span, the second term vanishes, establishing sufficiency. Otherwise, let $w=(I-\Pi_Q)q_*\ne0$. Then $w^\top Q=0$ and $w^\top q_*=\|w\|_2^2>0$. For any nonzero $d_v$-dimensional vector $u$, construct $S'=S+uw^\top$. These two states satisfy
\begin{equation}
  \begin{aligned}
    S'Q &= SQ,\\
    S'q_*-Sq_* &= u\|w\|_2^2\ne0.
  \end{aligned}
\end{equation}
They agree on all exposed readouts but differ on the new one, proving necessity.

The proposition concerns multiple queries to the same fixed state. Actual target features are formed at different positions and layers and include residual connections and nonlinear transformations. It therefore does not establish that the complete feature interface necessarily loses information relevant to drafting. Even when that interface determines the state, direct state access can change the computation required of a finite-capacity drafter. Whether this access improves prediction is an empirical question.

\subsection{Prefix Contributions to Future Readouts}
\label{app:future-readout-decomposition}

For Gated DeltaNet~\citep{yang2025gateddeltanet}, the update coefficients are
\begin{equation}
  \begin{aligned}
    A_i &= \alpha_i(I-\beta_i k_i k_i^\top),\\
    B_i &= \beta_i v_i k_i^\top,
  \end{aligned}
\end{equation}
so that $S_i=S_{i-1}A_i+B_i$. Fix a realized target-model continuation and its associated coefficients and queries. Define ordered products and accumulated writes by
\begin{equation}
  \begin{aligned}
    P_j &= A_{t+1}\cdots A_{t+j},\\
    W_j &= \sum_{a=1}^{j} B_{t+a}
      \bigl(A_{t+a+1}\cdots A_{t+j}\bigr),
  \end{aligned}
\end{equation}
where an empty product is the identity matrix.

\paragraph{Proposition 2.}
For every $j\ge1$, $S_{t+j}=S_tP_j+W_j$. Consequently, the raw attention readout satisfies Equation~\eqref{eq:future-state-readout}.

\paragraph{Proof.}
The claim holds for $j=1$ by the update rule. Assuming it holds for $j$, one further update gives
\begin{equation}
  \begin{aligned}
    S_{t+j+1}
      &= (S_tP_j+W_j)A_{t+j+1}+B_{t+j+1}\\
      &= S_tP_{j+1}+W_{j+1},
  \end{aligned}
\end{equation}
which proves the state identity by induction. Right-multiplication by $q_{t+j}$ yields the readout decomposition.

This is an algebraic decomposition along a fixed continuation. Future coefficients and queries depend on that continuation, so $q_j^{\mathrm{eff}}=P_jq_{t+j}$ is not an input available to the drafter. Nor does the first term recover subsequent writes, $W_jq_{t+j}$. ReaDy instead learns position-conditioned queries through the drafting objective, with normalization and low-rank projections restricting the practical reader. The analysis motivates an additional route to committed-prefix context; it does not assert exact recovery of future readouts.

%% file: ready-training-appendix.tex
\section{Training Details}
\label{app:training}

\paragraph{Training data.}
Open-PerfectBlend is a public instruction mixture covering mathematics, code, conversation, and instruction following. Its eight source corpora are listed in Table~\ref{tab:training-mixture}, using counts reported by the dataset card. The card reports removing 88.1k examples during deduplication. These counts describe the public mixture, rather than the source proportions of our 20,000- and 50,000-sample adaptation sets.

\begin{table}[!htb]
  \centering
  \caption{Source-corpus counts reported by the Open-PerfectBlend dataset card, before mixture deduplication. These are not per-source counts for ReaDy's training subsets.}
  \label{tab:training-mixture}
  \small
  \setlength{\tabcolsep}{4pt}
  \renewcommand{\arraystretch}{1.03}
  \begin{tabular*}{\columnwidth}{@{\extracolsep{\fill}}lr@{}}
    \toprule
    Source corpus & Reported samples \\
    \midrule
    MetaMathQA & 395,000 \\
    UltraInteract SFT & 288,579 \\
    UltraChat & 207,865 \\
    Orca-Math & 200,035 \\
    UltraFeedback & 187,405 \\
    Evol-CodeAlpaca & 111,272 \\
    AutoIF & 61,492 \\
    LMSYS Arena human preferences & 57,362 \\
    \bottomrule
  \end{tabular*}
\end{table}

\paragraph{Training mode.}
In its offline experiments, the original DSpark paper~\citep{cheng2026dspark} regenerates responses to Open-PerfectBlend prompts using each target model and adopts non-thinking mode for data generation and evaluation. Our two-stage adaptation instead targets thinking-mode generation, consistent with our evaluation setting. The non-thinking setting here refers to the original paper, not an assumption about the training data of the RadixArk checkpoint.

\paragraph{Loss aggregation.}
Each training anchor predicts seven tokens. We weight draft position $j$ by $w_j=\exp(-j/4)$ for $j=0,\ldots,6$. Let $\mathcal{V}_i$ contain the valid anchor--position pairs $(a,j)$ in sequence $i$. With $\ell_{\mathrm{CE}}^{i,a,j}$ and $\ell_{\mathrm{L1}}^{i,a,j}$ denoting the corresponding cross-entropy and probability-distribution L1 losses, the full-sweep sequence loss is
\begin{equation*}
  \mathcal{L}_i =
  \frac{\sum_{(a,j)\in\mathcal{V}_i}w_j
  \bigl(\ell_{\mathrm{CE}}^{i,a,j}+0.1\ell_{\mathrm{L1}}^{i,a,j}\bigr)}
  {\sum_{(a,j)\in\mathcal{V}_i}w_j}.
\end{equation*}
We first normalize within each sequence, as above, then average the sequence losses across the four training GPUs. The confidence-loss weight is zero. We use no additional KL-divergence, feature-reconstruction, or reward loss.

\paragraph{Relation to DSpark training.}
We reuse a pretrained DSpark drafter, but do not reproduce its original training recipe. The published DSpark objective combines cross-entropy, probability-distribution L1, and confidence losses with weights 0.1, 0.9, and 1.0, respectively~\citep{cheng2026dspark}. Our adaptation instead uses Equation~\eqref{eq:adaptation-loss}, with the position weighting and aggregation described above. Thus, the original-checkpoint comparison measures the complete adaptation procedure, while the no-state fine-tuning comparison in Table~\ref{tab:finetuning-control} provides a separate reference for assessing the state adapter.

\paragraph{Optimization.}
Both stages use AdamW with FP32 master parameters and optimizer states. The peak learning rate is $1\times10^{-4}$ in the reader-only stage and $5\times10^{-5}$ in the joint-adaptation stage. Both stages use cosine decay after warmup. Table~\ref{tab:training-settings} lists the optimizer settings.

\begin{table}[!htb]
  \centering
  \caption{Optimizer settings for the two adaptation stages.}
  \label{tab:training-settings}
  \small
  \setlength{\tabcolsep}{4pt}
  \renewcommand{\arraystretch}{1.05}
  \begin{tabular*}{\columnwidth}{@{\extracolsep{\fill}}lr@{}}
    \toprule
    Setting & Value \\
    \midrule
    Optimizer & AdamW \\
    $(\beta_1,\beta_2)$ & $(0.9,0.999)$ \\
    $\epsilon$ & $10^{-8}$ \\
    Peak learning rate (stage 1) & $1\times10^{-4}$ \\
    Peak learning rate (stage 2) & $5\times10^{-5}$ \\
    Weight decay & 0 \\
    Gradient clipping threshold & 1.0 \\
    \bottomrule
  \end{tabular*}
\end{table}

%% file: ready-qualitative-figure.tex
\begin{figure*}[!t]
  \centering
  \includegraphics[width=\textwidth]{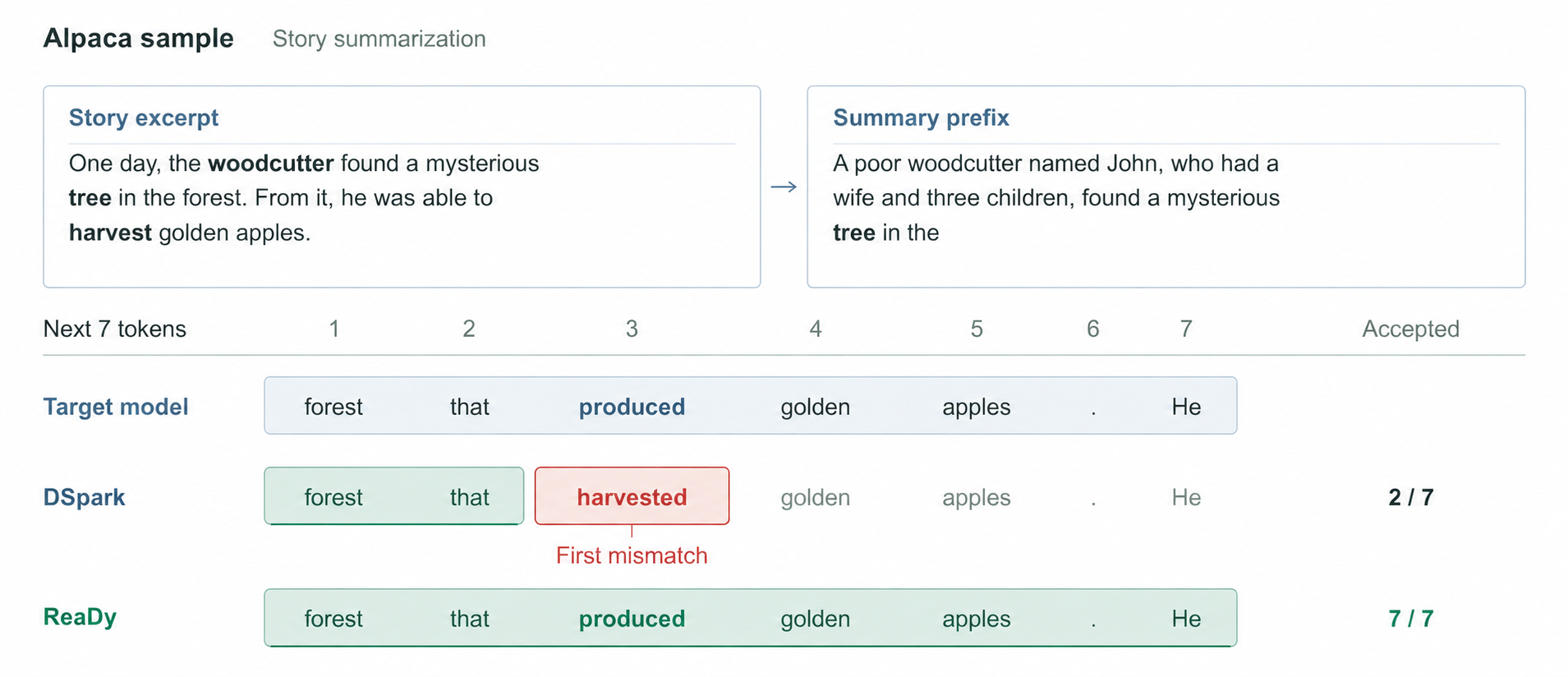}
  \caption{An Alpaca story-summarization example. Green highlights accepted proposed tokens; red marks the first mismatch. DSpark matches two of the seven proposals before its first mismatch, while ReaDy matches all seven. These counts refer only to accepted draft proposals, excluding target-supplied tokens.}
  \label{fig:ready-qualitative-example}
\end{figure*}

%% file: ready-qualitative-example.tex
\section{Qualitative Example}
\label{app:qualitative-example}

Figure~\ref{fig:ready-qualitative-example} compares seven-token proposals from DSpark and ReaDy at the same summary prefix in an Alpaca story-summarization example. DSpark first differs from the target at the third position, proposing ``harvested'' where the target produces ``produced.'' This mismatch ends its accepted prefix after two proposed tokens, even though its remaining displayed tokens match the target continuation. ReaDy matches all seven target tokens. The example illustrates how agreement at an early draft position determines how much of the proposed block can be accepted.

%% file: ready-background-appendix.tex
\section{Background}
\label{app:backbone-background}
\setcounter{equation}{0}
\renewcommand{\theequation}{\Alph{section}.\arabic{equation}}

\subsection{Linear and Hybrid Linear Attention}
\label{app:attention-background}

During autoregressive decoding, full attention reads token-wise keys and values whose storage grows with the prefix. Kernel linear attention instead aggregates past key--value associations into a recurrent state~\citep{katharopoulos2020linear}. For a positive feature map $\phi$, a causal attention head can be written as
\begin{equation}
  \begin{aligned}
    S_t &= S_{t-1}+v_t\phi(k_t)^\top,\\
    z_t &= z_{t-1}+\phi(k_t),\\
    o_t &= \frac{S_t\phi(q_t)}{z_t^\top\phi(q_t)},
  \end{aligned}
  \label{eq:kernel-linear-attention}
\end{equation}
with zero initial states. Here $q_t$, $k_t$, and $v_t$ are the current query, key, and value; $z_t$ supplies the normalization. At fixed feature dimensions, the state size and the cost of one recurrent update and readout are independent of prefix length. The state stores aggregated associations, whereas the query selects a particular readout.

Subsequent designs improve how this limited memory is updated. Data-dependent gates control retention, while delta-rule updates revise the association addressed by the current key. Gated DeltaNet combines these mechanisms~\citep{yang2025gateddeltanet}. For one head with normalized keys, its update can be written as
\begin{equation}
  \begin{aligned}
    \widetilde S_t &= \alpha_t S_{t-1},\\
    e_t &= v_t-\widetilde S_t k_t,\\
    S_t &= \widetilde S_t+\beta_t e_t k_t^\top,\\
    o_t &= S_tq_t.
  \end{aligned}
  \label{eq:gated-delta-update}
\end{equation}
In the original parameterization, the input-dependent gates $\alpha_t,\beta_t\in(0,1)$ control retention and write strength, respectively. The decay gate first attenuates the previous state. The delta term then corrects the difference between the incoming value $v_t$ and the value retrieved by $k_t$ from that decayed state. This supports both global forgetting and a targeted update to the current key--value association. Here $o_t$ denotes the raw attention readout, before the block's output normalization and gating; the kernel-attention denominator in Equation~\ref{eq:kernel-linear-attention} is not used.

Expanding Equation~\ref{eq:gated-delta-update} gives
\begin{equation}
  \begin{aligned}
    S_t &= S_{t-1}A_t+B_t,\\
    A_t &= \alpha_t(I-\beta_t k_tk_t^\top),\\
    B_t &= \beta_t v_tk_t^\top.
  \end{aligned}
  \label{eq:gated-delta-transition}
\end{equation}
Thus, $A_t$ transforms existing associations and $B_t$ writes the new value. This is the state transition used in Section~\ref{sec:theoretical-motivation} to separate the contribution of the committed prefix from subsequent writes.

Hybrid linear attention interleaves recurrent layers with full-attention layers, combining compact state-based context modeling with access to individual cached tokens~\citep{qwen2025next}. The recurrent layers maintain fixed-size states, while full-attention layers retain growing KV caches. This layer organization also shapes activation dynamics: massive activations form spikes before full-attention layers and can persist as plateaus across intervening linear-attention layers~\citep{su2026massive}. ReaDy accesses the recurrent states as supplementary context for drafting, without changing the target's hybrid architecture.

\subsection{Speculative Decoding}
\label{app:speculative-decoding}

In standard speculative decoding~\citep{leviathan2023speculative}, an autoregressive drafter samples a block of $b$ candidates $y_1,\ldots,y_b$. Let $q_j$ and $p_j$ denote the draft and target distributions conditioned on the committed prefix and $y_{<j}$. The target evaluates these prefixes in parallel. Candidates are checked from left to right, accepting $y_j\sim q_j$ with probability
\begin{equation}
  \alpha_j(y_j)=\min\!\left(1,\frac{p_j(y_j)}{q_j(y_j)}\right).
  \label{eq:speculative-acceptance}
\end{equation}
At the first rejection, the remaining proposals are discarded and a replacement is sampled from
\begin{equation}
  p_j^{\mathrm{corr}}(v)
  =\frac{[p_j(v)-q_j(v)]_+}
  {\sum_{u\in\mathcal V}[p_j(u)-q_j(u)]_+},
  \label{eq:speculative-correction}
\end{equation}
where $[x]_+=\max(x,0)$. If all candidates are accepted, an additional token is sampled from $p_{b+1}$. This procedure preserves the target distribution.

Our greedy evaluation accepts the longest prefix matching the target's argmax predictions. The first mismatch ends the accepted prefix. Our reported $\tau$ counts all committed tokens per verification, including target-supplied correction or bonus tokens, rather than only accepted proposals. Except when generation terminates, accepting $a$ proposals yields $a+1$ committed tokens.

\subsection{Target and Drafter Architectures}

\paragraph{Hybrid target model.}
Qwen3.8-27B has a dense language backbone with 64 layers and hidden width 5,120.\footnote{\url{https://huggingface.co/Qwen/Qwen3.8-27B}} Its layer pattern repeats three Gated DeltaNet layers followed by one gated full-attention layer, yielding 48 recurrent layers and 16 full-attention layers. The former retain fixed-size matrix states, while the latter use token-wise KV caches. The published configuration specifies 128-dimensional recurrent keys and values, giving the $128\times128$ per-head states accessed by ReaDy. Under zero-based indexing, the selected source layers 30 and 62 are both recurrent layers. ReaDy leaves this hybrid target architecture unchanged and adds a read-only interface to its committed recurrent states.

\paragraph{DSpark architecture.}
DSpark~\citep{cheng2026dspark} combines a DFlash-based parallel backbone~\citep{chen2026dflash} with a lightweight sequential prediction head. The backbone processes a block in one forward pass to produce position-wise representations and base logits. A sequential head adds transition biases conditioned on earlier proposals: the paper's Markov variant uses the preceding token through a low-rank transition map, while its RNN variant carries a within-block hidden state. DSpark also introduces confidence-based verification scheduling, separately from this drafting architecture. ReaDy supplies additional target-state context to the backbone through residual updates.

\paragraph{Checkpoint used in this work.}
The RadixArk Qwen3.8-27B-DSpark model card describes five full-attention draft layers with hidden width 5,120, 32 query heads, and eight KV heads. It receives auxiliary target features from layers 5, 19, 33, 47, and 61 and uses a rank-256 Markov head. These are properties of the released drafter that initializes ReaDy; the rank-eight setting in Section~\ref{sec:state-context} belongs to our state reader. The original hidden-feature path and the added recurrent-state path therefore expose different interfaces to target context.

%% file: ready-position-protocol.tex
\section{Draft-Prefix Survival Protocol}
\label{app:position-protocol}
\setcounter{equation}{0}
\renewcommand{\theequation}{\Alph{section}.\arabic{equation}}

Both methods decode identical prompts with greedy generation, concurrency one, a maximum of 1,024 new tokens, and natural EOS termination. Diagnostic tracing and uninstrumented runs produce identical output token sequences. We pool fully observable drafting rounds separately for each method and benchmark. For $N$ eligible rounds, let $A_i$ denote the number of accepted proposals in round $i$. For $j=1,\ldots,7$, the plotted estimator is
\begin{equation}
  \widehat{P}_j=\frac{1}{N}\sum_{i=1}^{N}
  \mathbf{1}\{A_i\ge j\}.
  \label{eq:prefix-survival-estimator}
\end{equation}
Terminal rounds with incomplete observations are censored, rather than counted as rejection events. The methods have different drafting boundaries, so their curves aggregate different rounds along identical output sequences.